# Bias Analysis of AI Models for Undergraduate Student Admissions

Kelly Van Busum*
Department of Computer Science and Software Engineering
Butler University
Indianapolis, Indiana, United States
kvanbusum@butler.edu
ORCID:  0000-0002-9443-084X

Shiaofen Fang
Indiana University Luddy School of Informatics, Computing and Engineering
Indiana University Indianapolis
Indianapolis, Indiana, United States
shfang@iu.edu
ORCID:  0000-0001-8277-5202

**ABSTRACT**

Bias detection and mitigation is an active area of research in machine learning.  This work extends previous research done by the authors [1] to provide a rigorous and more complete analysis of the bias found in AI predictive models.  Admissions data spanning six years was used to create an AI model to determine whether a given student would be directly admitted into the School of Science under various scenarios at a large urban research university.  During this time, submission of standardized test scores as part of a student's application became optional which led to interesting questions about the impact of standardized test scores on admission decisions.  We developed and analyzed AI models to understand which variables are important in admissions decisions, and how the decision to exclude test scores affects the demographics of the students who are admitted. We then evaluated the predictive models to detect and analyze biases these models may carry with respect to three variables chosen to represent sensitive populations: gender, race, and whether a student was the first in his/her family to attend college.  We also extended our analysis to show that the biases detected were persistent. Finally, we included several fairness metrics in our analysis and discussed the uses and limitations of these metrics.

**STATEMENTS AND DECLARATIONS**

- Competing Interests:  The authors have no relevant financial or non-financial interests to disclose.
- Ethics Approval:  This study was reviewed by the IUPUI Institutional Review Board.
- Data Availability:  The datasets generated during and/or analyzed during this study are not publicly available because they are the property of IUPUI.
- Author Contributions: Both authors contributed equally to all parts of this work.

**ACKNOWLEDGEMENTS**

The authors want to thank the following people at IUPUI for their help in providing the dataset used in this study, for answering questions about the dataset format, and for explaining policies related to admissions:  Jane Williams, Joe Thompson, Steve Graunke, Matt Moody, Norma Fewell, and Lori Hart.

# 1 INTRODUCTION

Artificial Intelligence (AI) has started to play an increasingly important role in higher education. AI-based predictive models built from existing datasets such as admissions data provide insight into the policies, practices, and biases in U.S. higher education system. At the same time, AI models have their own limitations and potential for harm such as biases towards certain populations. It is therefore important to understand these limitations and biases to truly integrate AI tools in future higher education applications. This paper describes an attempt to build, apply, and analyze AI models using an existing student admission dataset. An initial analysis of the AI models showed that the models contain bias, so a more rigorous and complete analysis of the bias was conducted.

Admissions data from the School of Science at a large urban research university was used to create machine learning-based AI models. These models predict whether a student would be directly admitted into the School of Science, or not, under a variety of scenarios. The dataset spans six years, and over this time, the admissions policy of the university changed from requiring students to submit standardized test scores as part of their application, to making test scores optional. Such fundamental changes in admission policies can significantly impact our higher education system for different populations, so it is important to have a strong understanding of the implications of such policy changes.

When existing data is used to create predictive models, the behaviors and features of these models provide a powerful opportunity to better understand both the potential impact of policies on different populations, and the limitations of the models themselves. This paper attempts to address both issues using a dataset collected from past admission data from the School of Science. Our analysis will first try to better understand what variables are important in admissions decisions, and how the decision to exclude test scores may affect the demographics of the students who are admitted. The predictive model contains a variety of demographic variables, and three: gender, race, and whether a student was the first in his/her family to attend college ("First-Generation"), were chosen to represent sensitive populations ("sensitive variables"). We then carefully evaluate the AI models for presence of potential biases with respect to performance relative to these three variables. The result of this analysis provides some initial evidence that AI algorithms can be harmful when used as part of the admission decision-making process if bias is not effectively mitigated.

This paper extends our earlier work [1] in this area in four ways. First, we used a larger dataset. This dataset contains an extra year of data, as well as some students who were excluded from the previous dataset. Next, in our previous work we presented an analysis for students admitted under the test-required policy and a separate analysis for those admitted under the test-optional policy. In this paper, we redid that analysis using the updated dataset, and added a third analysis where we analyzed the entire dataset with test scores removed. This was to simulate a larger test-optional dataset. Third, we conducted a more in-depth analysis of the bias found in the AI models to ensure that it persists. Finally, we include a discussion of additional metrics for fairness so that our bias detection is more rigorous and complete.

Students who attend large urban universities have some unique challenges. They may be more likely than students who attend other types of universities to face the competing demands of work, family, and school leading to poorer academic experiences. Attending a university in a city with a high cost-of-living and the continuing effects of COVID-19 are both factors that can increase the financial demands on students, also leading to increased stress and poorer academic outcomes. Although urban universities tend to be more diverse, minority students still report struggling with a sense of belonging, which can negatively affect academic performance. To best support all students, care must be taken to understand the impact of policy on equity.

Test-optional policies were designed to address the concern that standardized test scores are biased metrics for predicting student success and to increase equity in admissions procedures. But more research needs to be done to fully understand how these policies might change the demographics of the students admitted to the university and other impacts. Identifying the important factors in admissions decisions and how test-optional policies might change admitted student demographics is the first goal of this project.

As an additional step toward understanding and mitigating bias in admissions policies, college admissions officers are increasingly turning to the use of machine learning-based AI algorithms. However, when AI models are trained using existing datasets, the models can introduce new bias and fairness issues with the potential for harm. This shows a need to better understand how bias might manifest as AI algorithms become more widely used in higher education. Identifying and measuring bias in the models built to predict admission decisions is another important goal of this project.

This work makes the following contributions:

1. College admissions data was used to generate AI predictive models, evaluated on their overall accuracy and effectiveness. These models were used to better understand the primary factors in admissions decisions and their variations within different cohorts.
2. The performance of the AI predictive models was evaluated relative to various sensitive groups, and the results show evidence of biases. This serves as a warning and contributes to an understanding of how bias can be identified, measured, and eventually mitigated.
3. Fairness metrics were included in the analysis of the models, illustrating some limitations with the use of these metrics.

The rest of this paper is organized as follows. In Section 2, we provide an overview of the literature related to this work and its impact. In Section 3, we discuss the methods used to analyze the admissions dataset, the details of the AI prediction model, and metrics related to fairness and bias. In Section 4, we describe the results of our analysis and identify the bias found in the AI predictive model. In Section 5 we provide some concluding remarks and directions for future research.

## 2 RELATED WORK

Students at urban universities face some unique challenges. For example, they are likely to experience the conflicting demands of work, family, and school, which can negatively affect their satisfaction with educational experiences 0. Students, especially minority students, may face barriers that undermine academic achievement, reduce their sense of belonging, and interfere with degree completion 0. Those who attend a university located in a city with a lack of affordable housing may struggle to find appropriate housing which can negatively affect their academic lives, health, and well-being 0. Data collected about the impact of COVID-19 on urban college students showed that decreases in student earnings and household incomes led to significant disruptions in students' lives, and that these disruptions had an especially negative impact on first-generation students 0.

Recently, many universities have started to experiment with an admissions policy that allows students to apply without submitting standardized test scores. The hope is that these policies will address concerns of bias in the use of test scores as a metric to predict student success. One survey suggests that test-optional policies are changing enrollment demographics, especially with respect to underrepresented minorities, as Black and Hispanic students are 24% and 21% respectively more likely to apply to a school with a test-optional policy [6]. A second study found that test-optional admission increased the first-time enrollment by 10-12% of underrepresented minorities, and 6-8% by women 0. Another study of liberal arts schools found that although test-optional policies enhance the perceived selectivity of a school, they did not increase the diversity 0. Researchers are also exploring nuances of the policy. For example, what are the implications of giving students the option to submit test scores, when they are not required? Does it matter that some students choose not to submit scores because they are too low, while others choose not to take the test at all 0?

Artificial intelligence tools, particularly machine learning algorithms, are increasingly used in higher education applications. Using AI to assist college admission process can be more objective and efficient 0. Machine learning algorithms can also help universities better understand admission criteria and their impact in the admissions process 0 as well as admissions yield 0. AI can also be used to predict the likelihood of admission for individual students [13-14] in some situations and can sometimes provide evidence for bias in human-centered admissions processes 0.

While AI tools bring many benefits to higher education applications, they can also be biased towards sensitive populations due to the intrinsic bias in existing datasets and in the algorithms themselves 0. There are a variety of frameworks and taxonomies for classifying bias 0 and although some bias can be neutral/unobjectionable, many biases are problematic and require a response 0. Identifying and categorizing bias in AI is a first step toward creating methods for mitigating bias, an area of active research [19-20].

The social implications of the use of AI in education are nuanced, providing further motivation for careful understanding of the effects of algorithmic bias. For example, student perceptions of fairness with respect to the use of AI in college admissions can affect organizational reputation and the likelihood of students leaving the university [21]. Algorithmic bias can also affect the degree to which people are comfortable accepting AI-based recommendations or adopting AI systems [22]. Complicating this understanding is the gap in the literature between technical studies (which include jargon) and descriptive studies (aimed at the lay-person) on algorithmic fairness in higher education [23-24], showing a need for scholarly work which bridges these disciplines.

## 3  METHODS and DATASET

This research was conducted using admissions data from the School of Science at a large urban research university. Students in the dataset were applying for admission from Fall 2017- Spring 2023. The admissions process became test-optional during this period; first-year students applying for admission from Spring 2021 forward could choose not to submit standardized test scores. We built various machine learning-based models using different training sets and feature sets to predict which students were admitted directly into the School of Science ("Direct Admits"). Students who are categorized as "Not Direct Admits" are those who were admitted into the university but not into the School of Science, and those who were not admitted to the university at all.

Prior to test-optional admissions, Direct Admit decisions were based on grade point average (GPA) and standardized test scores. When the admissions process became test-optional, Direct Admit decisions for students who opted to exclude standardized test scores from their application were based on GPA and an assessment of "math readiness" based on performance in high-school math courses. The dataset used in this research contains approximately 11,600 students who were required to submit test scores, and approximately 7,900 students who were applying under the new test-optional policy.

The features used in the predictive model were: whether the student is a beginning student, the number of campuses to which the student applied, age, gender, race/ethnicity, whether the student is a first-generation student, whether the student is an in-state resident, GPA, and standardized test scores. Standardized test scores include scores from the two prevalent American college admissions tests, the ACT (American College Test) and SAT (formerly known as the Scholastic Assessment Test, but now the acronym does not stand for anything). ACT scores were normalized to SAT scores. In this dataset, ethnicity (Hispanic/Latino) was part of the race variable. Through exploratory analysis, three variables were chosen as variables to represent sensitive populations for further analysis: gender, race/ethnicity, and first-generation students.

This study consisted of an analysis of three groups of data:
- Group 1: Data over all six years.
- Group 2: Test-required cohort (Fall 2017 – Fall 2020).
- Group 3: Test-optional cohort (Spring 2021 – Spring 2023).

For each of these groups, we conducted three analyses:
- GPA was included, but standardized test scores were excluded.
- Standardized test scores were included, but GPA was excluded.
- Both GPA and test scores were included.

Finally, each analysis was repeated three times: once where the sensitive variable was gender, once where the sensitive variable was race, and once where the sensitive variable was first-generation students. The goal was to understand how changing the value of a single sensitive variable affected the accuracy of the models. A summary of the findings is illustrated in Figure 1. In addition, we note that the test-optional cohort contains only two-and-a-half years of data, and that recent data is confounded by the pandemic.

**Figure 1: Analysis Procedures**

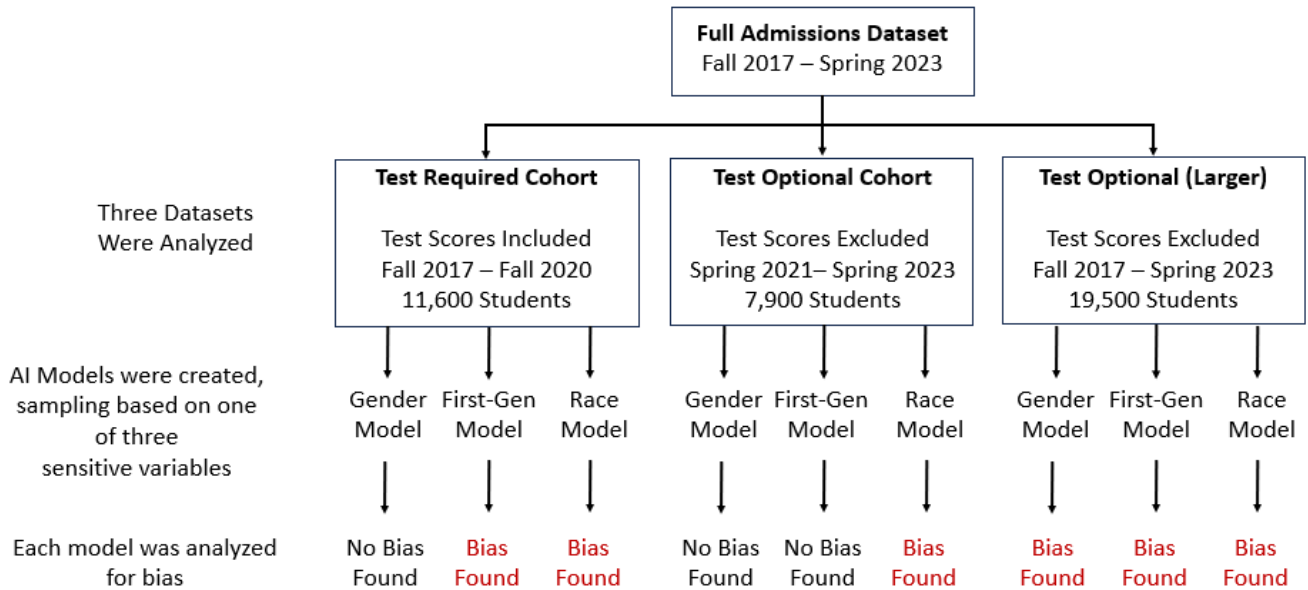

Demographic data with respect to the sensitive variables is summarized in Table 1. The demographic makeup of the applicants is heavily skewed towards white, female students whose parents also attended a university.

**Table 1: Sensitive Variables, Full Dataset**

| Gender | Race | First-Generation |
|---|---|---|
| Male (37%) | White (55%) | No (76%) |
| Female (63%) | Hispanic (13%) | Yes (24%) |
| | Black (12%) | |
| | Asian (7%) | |
| | Other (13%) | |

The predictive models were built using a linear support vector machine. Other machine learning models can also be applied here and may lead to better prediction accuracies, but that was beyond the scope of this research. We randomly selected two-thirds of the data for training each time, using proportionate stratified sampling with respect to the sensitive variable. All three of the sensitive variables were included in each model, but only one was treated as a sensitive variable for sampling in each experiment. Each model was then validated using five-fold cross-validation.

The resulting model was evaluated for potential bias, and to do this a bias metric needed to be identified for bias detection. As our main goal in this work was to identify bias in the AI models (as opposed to evaluating the fairness of admissions policies), we evaluate each model based on overall accuracy, as well as accuracies in each subgroup of the sensitive variables: Male/Female, White/Non-White, and Non-First-Generation/First-Generation. A difference in accuracy over a threshold of 5% between sub-groups was considered biased. The 5% threshold is selected somewhat subjectively and can certainly be modified based on the application needs. We also apply the same accuracy threshold to evaluate and compare the specificities and sensitivities of the model for different sub-groups. In this work, a lower Specificity score means a greater number of students incorrectly admitted. A lower Sensitivity score means a greater number of students incorrectly denied admission. The goal here was to experiment with an approach to helping people evaluate AI models trained from admission data, rather than determining if this particular dataset carries bias or not. Different threshold values may be used for different applications or scenarios.

Biases detected in Specificity and Sensitivity often lead to real fairness issues in the admission process, so we also included three fairness metrics: Brier Score, Balance for the Negative Class, and Balance for the Positive Class. A Brier score is the measure of agreement between the ground truth data and the predicted data; a lower Brier Score indicates that the model is highly calibrated. In the work, Balance for the Negative Class means that the average probability score assigned to students the model predicts will not be admitted should be the same across the subgroups of a sensitive variable. Similarly, Balance for the Positive Class means that the average probability score assigned to students the model predicts will be admitted should be the same across subgroups of a sensitive variable. These three fairness metrics cannot be satisfied simultaneously without perfect prediction or equal base rates, leading to conflicts between different definitions of fairness [25]. Note that fairness is a broader concept than bias. For example, admission policies and practices can be unfair by these definitions, but that does not imply bias in the AI algorithm itself.

## 4   RESULTS AND DISCUSSION

The results presented here are based on the accuracy of the AI predictive models in various scenarios. An exploratory analysis of the data and corresponding AI models found that:
1. Standardized test scores play a major role in admissions decisions.
2. Gender and First-Generation are two other variables that play prominent roles in admissions decisions.
3. Many students who would not be admitted under test-required policies would be admitted under test-optional policies, including more students from sensitive populations.

We also evaluated the models for bias with respect to sensitive variables. The results were:
1. Gender: The model for the larger test-optional dataset predicts women will be incorrectly admitted more than men (Specificity bias).
2. First-Generation: The model for the test-required cohort and the larger test-optional dataset predicts that non-first-generation students will be incorrectly admitted more than first-generation students (Specificity bias), and that first-generation students will be incorrectly rejected more than non-first-generation students (Sensitivity bias). In addition, for the larger test-optional dataset, there is a discrepancy in the average predicted probability for admission (Balance for the Positive Class).
3. Race/Ethnicity: All three models predict that white students will be incorrectly admitted more than non-white students (Specificity bias). The model for the test-required cohort and the model for the larger test-optional both predict that non-white students will be incorrectly rejected more than white students (Sensitivity bias). There is also a discrepancy between the average predicted probability for rejection (Balance for the Negative class).

### 4.1 General Analysis
The general analysis focused on three scenarios:
1.  The test-required cohort (includes both GPA and standardized test scores; 11,600 students)
2.  The test-optional cohort (includes GPA but excludes test scores; 7,900 students)
3.  The full dataset (includes GPA, but excludes test scores; 19,500 students)

Scenarios 1 and 2 were chosen because they are real-life scenarios. Scenario 3 was chosen to simulate a more robust test-optional dataset. The predictive accuracy for Scenario 3 is 80%, lower than for the other two scenarios (as expected), but still very good.

Table 2 shows the permutation importance of the top ten variables included in the experiment.

**Table 2: Permutation Importance**

| Test-Required | Test-Optional | Test-Optional (Large) |
|---:|---:|---:|
| Test Scores .283 | GPA .277 | GPA .248 |
| GPA .055 | Male .060 | Female .054 |
| Female .048 | Female .060 | First-Gen .043 |
| First-Gen .037 | Out-Of-State Res .053 | In-State Res .028 |
| Not First-Gen .035 | Not First-Gen .044 | Male .025 |
| Male .034 | First-Gen .042 | Out-of-State Res .025 |
| Out-Of-State Res .021 | In-State Res .036 | Not First-Gen .019 |
| In-State Res .021 | Black .010 | Black .011 |
| Black .003 | White .007 | Beginning Student .004 |
| Other Race .001 | Beginning Student .006 | Hispanic .002 |
| Cross Val: .88 (+/- .01) | .89 (+/- .02) | .80 (+/- .02) |

From this table, it is clear that test scores are the dominant variable for predicting Direct Admits in the test-required cohort and are well above the importance of GPA. For the test-optional cohort, GPA is the most important variable by far, as expected. Gender and First-Generation are other important variables in both models. It is somewhat surprising that, for the test-required cohort, test scores played a pivotal role in admissions, and GPA only played a minor role. It means that the decision of making test scores optional is perhaps more significant than many people believed, particularly for major urban public universities.

To better understand how the test-optional policy might change which students are admitted, in Figure 2 we plotted the values of GPA and Test Scores of all students in the test-required cohort who were not Direct Admits. Under the test-required policy, students with Test Scores under 1080 or a GPA under 3.0 would not be Direct Admits; these are the students in all but the upper-right quadrant. (Note that students in the upper-right quadrant are those whose test scores and grades qualify them to be Direct Admits. They were not Direct Admits because of incomplete applications, disciplinary concerns, or other extenuating circumstances.) The test-optional policy requires a higher GPA of 3.3 to be a Direct Admit, but many of those in the lower right quadrant clear that threshold and would become Direct Admits. These are the students whose GPAs are high enough under the test-optional policy to be Direct Admits, but whose Test Scores would exclude them under the test-required policy.

The significance of the admission policy change is evident in Figure 2 where many of the students who would not be directly admitted under a test-required policy may be admitted under a test-optional policy. This result is not a statement about which policy is better. It simply shows that a change from test-required to test-optional can fundamentally change the admissions decisions for a large number of students.

Table 3 shows how students in sensitive populations would be affected by the change from a test-required to a test-optional policy. More women, non-white students, and first-generation students would meet admissions thresholds under a test-optional policy. This data suggests that not only may there be more students admitted overall, but many of those students are also students who are part of a population more likely to report negative academic experiences. More research needs to be done to understand how to better prepare and support incoming students so that they might be successful academically.

**Figure 2: GPA and Test scores for Test-Required Cohort who were NOT Direct Admits**

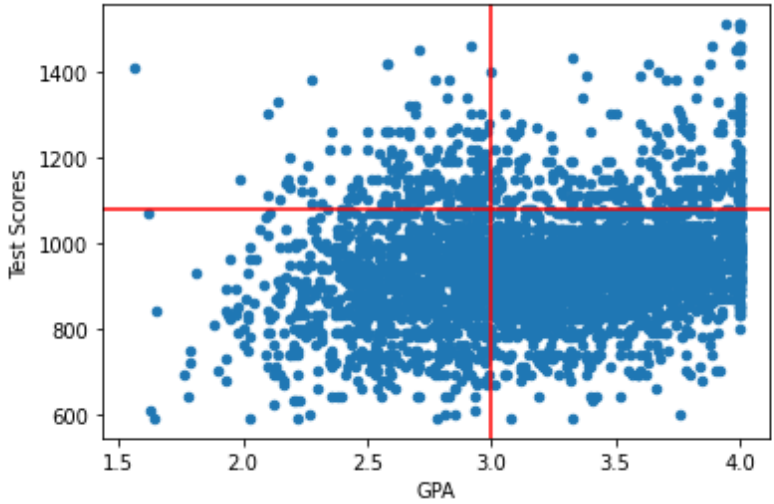

**Table 3: Demographics of Students in Test-Required Cohort Who Meet Admission Thresholds**

|  | Test-Required Threshold | Test-Optional Threshold |
|---|---|---|
| **Gender** | | |
| Female: | 61% | 68% |
| Male: | 39% | 32% |
| **Race** | | |
| Non-White: | 33% | 39% |
| White: | 67% | 61% |
| **First-Gen** | | |
| First-Gen: | 18% | 24% |
| Non-First-Gen: | 82% | 76% |

**4.2 Bias in Predictive Models**

In the process of identifying bias with respect to three sensitive variables in the predictive models, the focus was initially on two scenarios: the test-required cohort where the predictive model includes both GPA and test scores and the test-optional cohort where the predictive model excludes test-scores. Because the test-optional dataset only includes two-and-a-half years' worth of data, we added a third scenario where we create a predictive model of the entire dataset with test-scores excluded, to simulate a larger test-optional dataset. Note that this larger test-optional dataset may potentially have some demographic differences when compared to the true test-optional cohort of students, as simply removing test scores from the entire dataset does not account for the effects of a change in policy on the type of applicants it attracts. This dataset is intended to be used only to evaluate predictive models with a larger dataset and should not be used to infer policy implications.

Two of the sensitive variables are Gender and First-Generation, two variables with high permutation importance in both models. The third sensitive variable chosen was Race/Ethnicity, which was categorized as White and Non-White. The bias we tried to detect was a difference of 5% or more in overall prediction accuracy, or in Specificity or Sensitivity, between values of the sensitive variable. We did not find any significant differences in overall prediction accuracies between the sensitive variables for any of the three models. Differences in Specificity and/or Sensitivity were present in all models. In this application, Specificity is a measure of students who were incorrectly admitted, and sensitivity is a measure of students who were incorrectly denied admission.

The results for Gender are summarized in Table 4. None of the predictive models where Gender is the sensitive variable show bias.

**Table 4: Model Bias when Gender is Sensitive Variable**

|  | Overall | Male | Female |
|---|---|---|---|
| **Test-Required Cohort** | | | |
| Model Accuracy: | .872 | .875 | .870 |
| Specificity: | .841 | .843 | .841 |
| Sensitivity: | .893 | .893 | .893 |
| Cross Val: .88 (+/- .01) | | | |
| **Test-Optional Cohort** | | | |
| Model Accuracy: | .890 | .883 | .893 |
| Specificity: | .732 | .756 | .716 |
| Sensitivity: | .953 | .943 | .957 |
| Cross Val: .89 (+/- .02) | | | |
| **Test-Optional (Larger)** | | | |
| Model Accuracy: | .804 | .815 | .798 |
| Specificity: | .682 | .703 | .671 |
| Sensitivity: | .879 | .883 | .877 |
| Cross Val: .80 (+/- .02) | | | |

The results for First-Generation are shown in Table 5. The predictive models for both the test-required cohort and the larger test-optional dataset show bias. The larger test-optional model predicts that non-first-generation students are accepted more often than in reality (Specificity bias), and the models for both the test-required cohort and the larger test-optional dataset predict that first-generation students are rejected more often than in reality (Sensitivity bias).

**Table 5: Model Bias when First Gen is Sensitive Variable**

|  | Overall | Non-First-Gen | First-Gen |
|---|---|---|---|
| **Test-Required Cohort** | | | |
| Model Accuracy: | .886 | .890 | .873 |
| Specificity: | .855 | .841 | .884 |
| Sensitivity*: | .908 | .920 | .861 |
| Cross Val: .88 (+/- .01) | | | |
| **Test-Optional Cohort** | | | |
| Model Accuracy: | .881 | .889 | .851 |
| Specificity: | .727 | .733 | .709 |
| Sensitivity: | .941 | .947 | .918 |
| Cross Val: .89 (+/- .02) | | | |
| **Test-Optional (Larger)** | | | |
| Model Accuracy: | .807 | .817 | .772 |
| Specificity*: | .678 | .655 | .731 |
| Sensitivity*: | .885 | .904 | .810 |
| Cross Val: .80 (+/- .02) | | | |

The results for Race are presented in Table 6. All three models show bias when Race is the sensitive variable. All three models predict that white students will be incorrectly admitted more than non-white students (Specificity bias). The model for the test-required cohort and the model for the larger test-optional dataset show that non-white students will be incorrectly rejected more than white students (Sensitivity bias).

**Table 6: Model Bias when Race is Sensitive Variable**

|  | Overall | Non-White | White |
|---|---|---|---|
| **Test-Required Cohort** | | | |
| Model Accuracy: | .879 | .873 | .884 |
| Specificity*: | .854 | .882 | .824 |
| Sensitivity*: | .899 | .862 | .918 |
| Cross Val: .88 (+/- .01) | | | |
| **Test-Optional Cohort** | | | |
| Model Accuracy: | .885 | .871 | .896 |
| Specificity*: | .725 | .751 | .694 |
| Sensitivity: | .950 | .937 | .960 |
| Cross Val: .89 (+/- .02) | | | |
| **Test-Optional (Larger)** | | | |
| Model Accuracy*: | .804 | .775 | .826 |
| Specificity*: | .673 | .717 | .622 |
| Sensitivity*: | .883 | .826 | .917 |
| Cross Val: .80 (+/- .02) | | | |

It ought to be noted that biases identified here are based on ground truth data. They are not indications of any bias in the actual admission policies or practices. Rather it is an evaluation of errors present in the machine learning algorithms and how these errors were distributed unevenly for different student populations. These biases need to be understood if the models are to be used in the future for an AI-assisted admissions process, or other similar applications.

### 4.3 Aggregate Bias

It is important to note that the results presented above in Section 4.2 are a snapshot. For example, to examine bias in the test-required cohort when Gender is the sensitive variable, the test-required cohort was divided into training and testing datasets *once* and a model was created *once*. However, the overall accuracy of the predictive model generated for the test-required and test-optional cohorts is approximately 88%, and the overall accuracy of the predictive model generated for the test-optional cohort is approximately 80%. This is very high, which means that the number of outliers is small. This means that the distribution of outliers in the training and testing datasets may have a meaningful impact on the bias observed in the results.

To better understand the effects of outliers on the observed bias, we conducted an analysis of the stability of the bias for each model to ensure that any bias observed would persist regardless of how the dataset under examination was divided into training and testing datasets. For each model, we ran repeated trials. For each trial, two-thirds of the data was randomly selected for training, using proportionate stratified sampling with respect to the sensitive variable. We recorded the mean values for Specificity and Sensitivity, as well as the mean values for the Brier Score, Balance for the Negative Class and Balance for the Positive Class.

The results for Gender are shown in Table 7. The model for the larger test-optional dataset predicts women will be incorrectly admitted more than men (Specificity bias).

**Table 7: Aggregate Bias when Gender is Sensitive Variable**

|  | 10 Trials | | | 100 Trials | | | 500 Trials | | |
|---|---|---|---|---|---|---|---|---|---|
|  | Male | Female | Difference | Male | Female | Difference | Male | Female | Difference |
| **Test-Required Cohort** | | | | | | | | | |
| Specificity: | .820 | .865 | .045 | .818 | .862 | .044 | .816 | .860 | .044 |
| Sensitivity: | .912 | .896 | .016 | .913 | .896 | .017 | .912 | .896 | .016 |
| Brier Score: | .093 | .085 | .008 | .094 | .086 | .008 | .095 | .086 | .009 |
| Balance for Neg Class: | .848 | .850 | .002 | .847 | .848 | .001 | .847 | .850 | .003 |
| Balance for Pos Class: | .886 | .868 | .018 | .886 | .868 | .018 | .886 | .869 | .017 |
| Test-Optional Cohort | | | | | | | | | |
| Specificity: | .752 | .720 | .032 | .754 | .720 | .034 | .756 | .719 | .037 |
| Sensitivity: | .938 | .959 | .021 | .940 | .958 | .018 | .939 | .959 | .020 |
| Brier Score: | .094 | .090 | .004 | .094 | .090 | .004 | .094 | .091 | .003 |
| Balance for Neg Class: | .808 | .814 | .006 | .813 | .810 | .003 | .812 | .810 | .002 |
| Balance for Pos Class: | .874 | .887 | .013 | .876 | .886 | .010 | .876 | .887 | .011 |
| Test-Optional (Larger) | | | | | | | | | |
| Specificity*: | .712 | .655 | .057 | .713 | .656 | .057 | .711 | .655 | .056 |
| Sensitivity: | .890 | .884 | .006 | .890 | .883 | .007 | .888 | .883 | .005 |
| Brier Score: | .128 | .141 | .013 | .127 | .142 | .015 | .127 | .142 | .015 |
| Balance for Neg Class: | .788 | .789 | .001 | .789 | .790 | .001 | .789 | .790 | .001 |
| Balance for Pos Class: | .828 | .809 | .019 | .830 | .810 | .020 | .830 | .810 | .020 |

The results for First Generation are shown in Table 8. The model for the test-required cohort and the larger test-optional dataset predicts that non-first-generation students will be incorrectly admitted more than first-generation students (Specificity bias), and that first-generation students will be incorrectly rejected more than non-first-generation students (Sensitivity bias). In addition, for the larger test-optional dataset, there is a discrepancy in the average predicted probability for admission (Balance for the Positive Class).

**Table 8: Aggregate Bias when First Gen is Sensitive Variable**

|   | 10 Trials | | | 100 Trials | | | 500 Trials | | |
|---|---|---|---|---|---|---|---|---|---|
|   | Non-FGen | FGen | Difference | Non-FGen | FGen | Difference | Non-FGen | FGen | Difference |
| Test-Required Cohort | | | | | | | | | |
| Specificity*: | .823 | .890 | .067 | .822 | .892 | .070 | .823 | .893 | .070 |
| Sensitivity*: | .919 | .855 | .064 | .916 | .855 | .061 | .916 | .852 | .064 |
| Brier Score: | .089 | .091 | .002 | .089 | .090 | .001 | .089 | .090 | .001 |
| Balance for Neg Class: | .845 | .860 | .015 | .844 | .860 | .016 | .843 | .860 | .017 |
| Balance for Pos Class: | .884 | .837 | .047 | .885 | .837 | .048 | .885 | .837 | .048 |
| Test-Optional Cohort | | | | | | | | | |
| Specificity: | .724 | .750 | .026 | .727 | .749 | .022 | .727 | .755 | .028 |
| Sensitivity: | .953 | .943 | .010 | .955 | .944 | .011 | .954 | .942 | .012 |
| Brier Score: | .091 | .102 | .011 | .090 | .100 | .010 | .090 | .099 | .009 |
| Balance for Neg Class: | .811 | .803 | .008 | .812 | .808 | .004 | .812 | .807 | .005 |
| Balance for Pos Class: | .887 | .865 | .022 | .888 | .863 | .025 | .888 | .863 | .025 |
| Test-Optional (Larger) | | | | | | | | | |
| Specificity*: | .655 | .709 | .054 | .658 | .713 | .055 | .658 | .710 | .052 |
| Sensitivity*: | .907 | .812 | .095 | .904 | .807 | .097 | .905 | .806 | .099 |
| Brier Score: | .130 | .157 | .027 | .131 | .156 | .025 | .131 | .157 | .026 |
| Balance for Neg Class: | .784 | .798 | .014 | .786 | .798 | .012 | .786 | .798 | .012 |
| Balance for Pos Class*: | .827 | .774 | .053 | .827 | .774 | .053 | .827 | .773 | .054 |

The results for Race are shown in Table 9. All three models predict that white students will be incorrectly admitted more than non-white students (Specificity bias). The model for the test-required cohort and the model for the larger test-optional both predict that non-white students will be incorrectly rejected more than white students (Sensitivity bias). There is also a discrepancy between the average predicted probability for rejection (Balance for the Negative class).

Table 9: Aggregate Bias when Race is Sensitive Variable

|  | 10 Trials | | | 100 Trials | | | 500 Trials | | |
| --- | --- | --- | --- | --- | --- | --- | --- | --- | --- |
|  | Non-White | White | Difference | Non-White | White | Difference | Non-White | White | Difference |
| Test-Required Cohort | | | | | | | | | |
| Specificity*: | .891 | .805 | .086 | .886 | .802 | .084 | .887 | .803 | .084 |
| Sensitivity*: | .861 | .925 | .064 | .864 | .923 | .059 | .865 | .921 | .056 |
| Brier Score: | .091 | .086 | .005 | .092 | .087 | .005 | .091 | .087 | .004 |
| Balance for Neg Class*: | .879 | .819 | .060 | .876 | .817 | .059 | .875 | .816 | .059 |
| Balance for Pos Class: | .861 | .885 | .024 | .859 | .884 | .025 | .858 | .884 | .026 |
| Test-Optional Cohort | | | | | | | | | |
| Specificity*: | .763 | .710 | .053 | .756 | .703 | .053 | .756 | .705 | .051 |
| Sensitivity: | .929 | .963 | .034 | .934 | .964 | .030 | .935 | .964 | .029 |
| Brier Score: | .102 | .082 | .020 | .102 | .084 | .018 | .102 | .084 | .018 |
| Balance for Neg Class: | .822 | .791 | .031 | .825 | .794 | .031 | .824 | .794 | .030 |
| Balance for Pos Class: | .873 | .893 | .020 | .870 | .892 | .022 | .870 | .892 | .022 |
| Test-Optional (Larger) | | | | | | | | | |
| Specificity*: | .717 | .625 | .092 | .716 | .627 | .089 | .716 | .626 | .090 |
| Sensitivity*: | .832 | .912 | .080 | .834 | .916 | .082 | .835 | .915 | .080 |
| Brier Score: | .151 | .126 | .025 | .151 | .125 | .026 | .151 | .125 | .026 |
| Balance for Neg Class: | .810 | .764 | .046 | .810 | .765 | .045 | .809 | .765 | .044 |
| Balance for Pos Class: | .787 | .834 | .047 | .787 | .835 | .048 | .787 | .834 | .047 |

With respect to our fairness metrics: Brier Score, Balance for the Negative Class, and Balance for the Positive Class, note that differences with respect to subgroups of a sensitive variable only exceeded the 5% threshold in two instances. This means that in most cases all three fairness metrics are being satisfied simultaneously, which is happening because the predictive model is highly accurate [25]. The fact that bias is seen in Specificity and Sensitivity scores, but not reflected in the fairness metrics underscores an important limitation of bias and fairness metrics and helps to motivate further study in this area.

## 5 CONCLUSIONS

This research analyzed admissions data at a large urban research university using machine learning-based AI models to predict whether a given student was directly admitted into the School of Science. The dataset is interesting because the university's admissions policies changed from test-required to test-optional during this period. This research contributes to a better understanding of the variables that play an important role in admissions decisions and shows that there is a significant change in admitted student body under a test-optional policy. In addition, the predictive models create new bias with respect to different populations under sensitive variables of gender, race, and first-generation. The conclusions we can draw from this study include:

- Test-optional policies may significantly alter the demographics of the students who are admitted. The impacts of this need additional study.
- AI models are increasingly used to create and understand policies in higher education, so it is critical to carefully analyze these models for errors and biases. Further, the bias may not be apparent when examining the overall accuracy of the model, so more thorough evaluations must be done.
- Although policies may reflect social bias, separate metrics for bias and fairness must be developed when evaluating AI models themselves. Additionally, while useful, there are limitations to bias and fairness metrics that need to be better understood.

One avenue for future work in this area is to analyze a more expansive dataset. This research focused on School of Science data, but are the results the same when applied to a university-wide dataset? In addition, the test-optional policy is recent, and this dataset only includes about two years of test-optional data. Continuing this analysis over the next few years could provide some interesting insight and more robust results. Similarly, test scores for the test-optional cohort were excluded entirely in this analysis. The overall accuracy of these models was high, so this was sufficient for the models, though not entirely reflective of reality. Under a test-optional policy, test-scores can be included in admissions decisions, if this is what a student prefers. A model that includes this nuance may lead to additional insights about the effects of test-optional policies.

A second area for future work is in bias mitigation. When bias in AI models is detected, how can it be corrected or mitigated? For example, can carefully designed adjustments to the training set improve bias and fairness metrics in the model? Additional questions include: Are there scenarios under which it should not be corrected? What is the interplay between social bias reflected in admissions policies, and bias in AI models? etc.

As AI algorithms become more widely used, they can lead to greater accuracy, consistency, time-savings, and understanding in many domains. AI is a powerful tool and must be used with the understanding that as it is being used to solve problems, it can also introduce new problems. Techniques for identifying, measuring, and mitigating these problems are critical.

REFERENCES

[1] Kelly Van Busum and Shiaofen Fang. 2023. Analysis of AI Models for Student Admissions: A Case Study. *Proceedings of the 38th ACM/SIGAPP Symposium on Applied Computing (SAC '23)*, March 2023, 17-22. DOI: https://doi.org/10.1145/3555776.3577743

[2] Leslie B. Hammer, Tenora D. Grigsby, and Steven Woods. 1998. The Conflicting Demands of Work, Family, and School Among Students at an Urban University. *The Journal of Psychology* 132, 1 (1998), 220-226. DOI: https://doi.org/10.1080/00223989809599161

[3] Shaun R. Harper, Edward J. Smith, and Charles H. F. Davis, III. 2018. A Critical Race Case Analysis of Black Undergraduate Student Success at an Urban University. *Urban Education* 53, 1 (2018), 3-25. DOI: https://doi.org/10.1177/0042085916668956

[4] Luisa Sotomayor, Derya Tarhan, Marcelo Vieta, Shelagh McCartney, and Aida Mas. 2022. When students are house-poor: Urban universities, student marginality, and the hidden curriculum of student housing. *Cities* 124 (May 2022), 103572. DOI: https://doi.org/10.1016/j.cities.2022.103572

[5] Nuria Rodriquez-Planas. 2022. Hitting where it hurts most: COVID-19 and low-income urban college students. *Econ Ed Rev* (Apr. 2022), 102233. DOI: https://doi.org/10.1016/j.econedurev.2022.102233

[6] Jon Edelman. 2022. Survey: Test-Optional is Appealing to Minority Students. Retrieved Oct. 12, 2022 from https://www.diverseeducation.com/students/article/15292751/survey-finds-testoptional-policies-a-significant-motivator-for-minority-college-applicants

[7] Christopher T. Bennett. 2022. Untested Admissions: Examining Changes in Application Behavior and Student Demographics Under Test-Optional Policies. *American Educational Research Journal* 59, 1 (Feb. 2022), 180-216. DOI: https://doi.org/10.3102/00028312211003526

[8] Andrew S. Belasco, Kelly O. Rosinger, and James C. Hearn. 2015. The Test-Optional Movement at America's Selective Liberal Arts Colleges: A Boon for Equity or Something Else? *Educational Evaluation and Policy Analysis* 37, 2 (June 2015), 206-223. DOI: https://doi.org/10.3102/0162373714537350

[9] Zhi Liu and Nikhil Garg. 2021. Test-optional Policies: Overcoming Strategic Behavior and Informational Gaps. In *Proceedings of the ACM Equity and Access in Algorithms, Mechanisms, and Optimization (EEAAMO '21),* Oct. 5-9, 2021, ACM Inc., New York, NY, 1-13. DOI: https://doi.org/10.1145/3465416.3483293

[10] Austin Waters and Risto Miikkulainen. 2014. GRADE: Machine Learning Support for Graduate Admissions. *AI Magazine* 35, 1 (Spring 2014), 64-75. DOI: https://doi.org/10.1609/aimag.v35i1.2504

[11] Nicholas T. Young and Marcos D. Caballero. 2019. Using machine learning to understand physics graduate school admissions. arXiv: 1907.01570. Retrieved from https://arxiv.org/abs/1907.01570

[12] Joseph Jamison. 2017. Applying Machine Learning to Predict Davidson College's Admissions Yield. In *Proceedings of the 2017 ACM SIGCSE Technical Symposium on Computer Science Education (SIGCSE '17)*, Mar. 8-11, 2017, Seattle, Washington. ACM Inc., New York, NY, 765-766. DOI: https://doi.org/10.1145/3017680.3022468

[13] Mohan S. Acharya, Asfia Armaan, and Aneeta S. Antony. A Comparison of Regression Models for Prediction of Graduate Admissions. In *2019 International Conference on Computational Intelligence in Data Science (ICCIDS)*, Feb. 21-23, 2019, Chennai, India. IEEE, 1-5. DOI: 10.1109/ICCIDS.2019.8862140

[14] Ch. V. Raghavendran, Ch. Pavan Venkata Vamsi, T. Veerraju, and Ravi Kishore Veluri. 2021. Predicting Student Admissions Rate into University Using Machine Learning Models. In: D. Bhattacharyya and N. Thirupathi Rao, *Machine Intelligence and Soft Computing*. Advances in Intelligent Systems and Computing, vol 1280. Springer, Singapore. DOI: https://doi.org/10.1007/978-981-15-9516-5_13

[15] A.J. Alvero, Noah Arthurs, Anthony Lising Antonio, Benjamin W. Domingue, Ben Gebre-Medhin, Sonia Giebel, and Mitchell L. Stevens. 2020. AI and Holistic Review: Informing Human Reading in College Admissions. In *Proceedings of the AAAI/ACM Conference on AI, Ethics, and Society (AIES '20)*, Feb. 7-9, 2020, New York, NY. ACM Inc., New York, NY, 200-206. DOI: https://doi.org/10.1145/3375627.3375871

[16] Barbara Martinez Neda, Yue Zeng, and Sergio-Gago-Masague. 2021. Using Machine Learning in Admissions: Reducing Human and Algorithmic Bias in the Selection Process. In *Proceedings of the 52nd ACM Technical Symposium on Computer Science Education (SIGCSE '21)*, Mar. 13-20, 2021. Virtual. ACM Inc., New York, NY, 1323. DOI: https://doi.org/10.1145/3408877.3439664

[17]     Brian d'Alessandro, Cathy O'Neil, and Tom LaGatta. 2017. Conscientious Classification: A Data Scientist's Guide to Discrimination-Aware Classification. *Big Data* 5, 2 (Jun. 2017), 120-134. DOI: http://doi.org/10.1089/big.2016.0048

[18]     David Danks and Alex John London. 2017. Algorithmic Bias in Autonomous Systems. In *Proceedings of the Twenty-Sixth International Joint Conference on Artificial Intelligence (IJCAI '17)*, Aug. 19-25, 2017. Melbourne, Australia, 4691-4697. DOI: https://doi.org/10.24963/ijcai.2017/654

[19]     Ninareh Mehrabi, Fred Morstatter, Nripsuta Saxena, Kristina Lerman, and Aram Galstyan. 2019. A Survey on Bias and Fairness in Machine Learning. arXiv: 1908.09635. Retrieved from https://arxiv.org/abs/1908.09635

[20]     Simon Caton and Christian Haas. 2020. Fairness in Machine Learning: A Survey. arXiv: 2010.04053. Retrieved from https://arxiv.org/abs/2010.04053

[21]     Frank Marcinkowski, Kimon Kieslich, Christopher Starke, and Marco Lünich. 2020. Implications of AI (un-)fairness in higher education admissions: the effects of perceived AI (un-)fairness on exit, voice and organizational reputation. In *Proceedings of the 2020 Conference on Fairness, Accountability, and Transparency (FAT* '20)*, Jan. 2020, 122-130. DOI: https://doi.org/10.1145/3351095.3372867

[22]     Nima Kordzadeh and Maryam Ghasemaghaei. 2021. Algorithmic Bias: review, synthesis, and future research directions. *European Journal of Information Systems* (Jun. 2021), 388-409. DOI: https://doi.org/10.1080/0960085X.2021.1927212

[23]     Bahar Memarian and Tenzin Doleck. 2023. Fairness, Accountability, Transparency, and Ethics (FATE) in Artificial Intelligence (AI) and higher education: A systematic review. *Computers and Education: Artificial Intelligence* 5 (2023). DOI: https://doi.org/10.1016/j.caeai.2023.100152

[24]     Lorenzo Belenguer. 2022. AI bias: exploring discriminatory algorithmic decision-making models and the application of possible machine-centric solutions adapted from the pharmaceutical industry. *AI Ethics* 2, 4 (2022), 771-787. DOI: https://doi.org/10.1007/s43681-022-00138-8

[25]     Jon Kleinberg, Sendhil Mullainathan, and Manish Raghavan. 2016. Inherent Trade-Offs in the Fair Determination of Risk Scores. arXiv: 1609.05807v1. Retrieved from https://arxiv.org/pdf/1609.05807v1.pdf